\pdfoutput=1
\relax
\documentclass[letterpaper]{article}
\usepackage{aaai19}
\usepackage{times}
\usepackage{helvet}
\usepackage{courier}
\usepackage{epsfig}
\usepackage{graphicx}
\graphicspath{{images/}}
\usepackage{amsmath}
\usepackage{amssymb}
\usepackage{dsfont}
\usepackage{tablefootnote}
\usepackage{multirow}
\usepackage{subcaption}
\usepackage[ruled]{algorithm2e}
\usepackage{xcolor}
\usepackage{amsthm}
\usepackage{array}

\begin{document}
%
\title{Adversarial Classifier for Imbalanced Problems}
 \author{Ehsan Montahaei\thanks{These authors contributed equally to this work, The author order was determined by chance.}, Mahsa Ghorbani\footnotemark[1], Mahdieh Soleymani Baghshah and Hamid R. Rabiee \\
 Sharif University of Technology \\  ehsan.montahaei@gmail.com, mghorbani@ce.sharif.edu,\{soleymani, rabiee\}@sharif.edu  }

\maketitle

\begin{abstract}

Adversarial approach has been widely used for data generation in the last few years. However, this approach has not been extensively utilized for classifier training. In this paper, we propose an adversarial framework for classifier training that can also handle imbalanced data. Indeed, a network is trained via an adversarial approach to give weights to samples of the majority class such that the obtained classification problem becomes more challenging for the discriminator and thus boosts its classification capability. In addition to the general imbalanced classification problems, the proposed method can also be used for problems such as graph representation learning in which it is desired to discriminate similar nodes from dissimilar nodes. Experimental results on imbalanced data classification and on the tasks like graph link prediction show the superiority of the proposed method compared to the state-of-the-art methods.

\end{abstract}

\section{Introduction}
Imbalanced data is prevalent in real-world classification problems. If training data are not balanced, most of existing learning algorithms produce inductive bias towards the frequent classes. Therefore, they result in poor minority class recognition performance while the accuracy on minority classes can be much important, e.g., in rare event discovery \cite{deep_imbalanced}. For deep learning models, influence of imbalanced data on representation learning can also deteriorate the performance on majority classes \cite{deep_over}. A simple but common option for imbalanced data classification is re-sampling the data before training that includes over-sampling minority classes, under-sampling majority classes, or combination of these approaches \cite{deep_imbalanced}. The second option is cost-sensitive learning that assigns higher cost of mis-classification to the minority class than to the majority \cite{deep_cvpr}. Despite the continuous development of learning methods in the last decades, learning from imbalanced data is still a challenging problem \cite{Bartosz}. 

Let us consider the problem of discriminating samples of the intended class that are rare (compared to other samples). One example of such problems is discriminating similar pairs of data from dissimilar ones. Similar pairs are usually rare compared to dissimilar ones. For example, the link prediction problem in the graph can be seen as discriminating pairs of connected nodes from disconnected ones. Similarly, we encounter this situation when needing to discriminate similar questions in the Community Question Answering (CQA) forums.

One of the well-known approaches to handle imbalanced data is down-sampling the training data of the majority class. However, random down-sampling may not work appropriately since the selected subset of samples may not be informative enough.
Let us consider a two-class classification task with rare positive samples compared to the negative ones. If we select uniformly random the negative samples among the training data of this class, they may be those samples that are far from the boundary and so they may not be such helpful in training a powerful classifier. This scenario happens especially when most of the negative samples are far from the boundary. For example, when we want to distinguish similar pairs of data from dissimilar ones, most of the dissimilar pairs are clearly different and only a small portion of them may be challenging and helpful for training.

\begin{figure}[!t]
 \begin{center}
 \includegraphics[width=1\columnwidth]{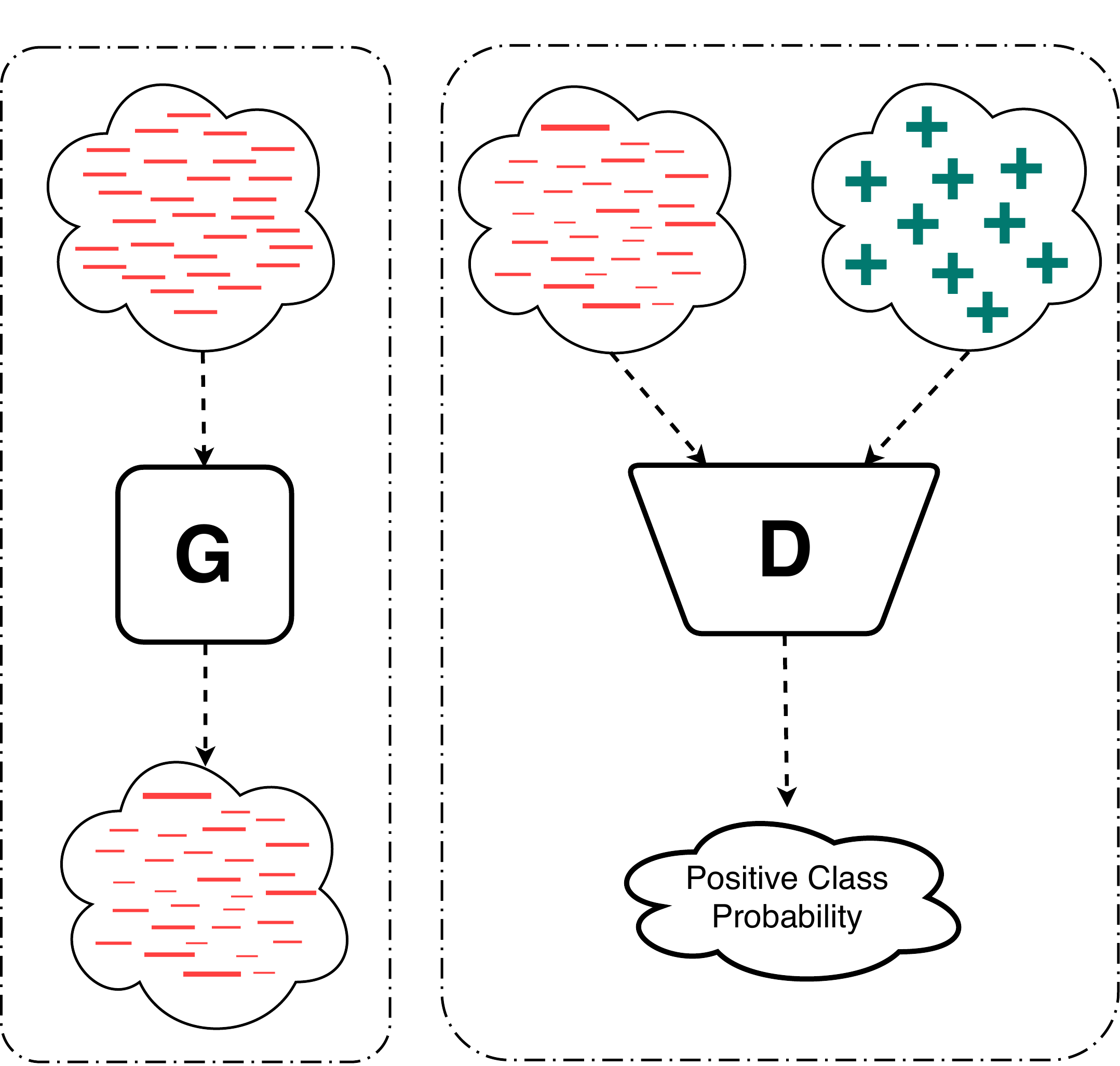}
 \caption{Illustration of the proposed adversarial classification framework. $G$ re-weights the negative samples and $D$ is trained using the positive and the re-weighted negative samples.}
 \label{fig:overview}
 \end{center}
 \end{figure}

In this paper, we propose an adversary-based approach to give weight to each sample of the majority class. The proposed method is called \textit{Adversarially Re-weighting for Imbalanced Classification} (ARIC). In fact, we learn a distribution on the majority class samples that indicates the probability of using corresponding samples. This distribution is achieved in an adversarial manner. The generator learns the distribution on samples of the majority class to provide a more challenging classification problem to the discriminator. On the other hand, the discriminator tries to accurately classify (the weighted) samples despite the generator has made the task more intricate (Fig.\ref{fig:overview}). At the end of this game, we intend to reach a more accomplished discriminator since it has been trained in an adversarial manner by focusing on more instructive samples of the majority class. The main contributions of our work are:
\begin{enumerate}
    \item Training classifiers via an adversarial approach for imbalanced classification problems. We propose a general framework for this problem and train a new type of generator that provides weights for samples of the majority class.
        \item Experiments were conducted to evaluate the proposed adversarial classifier on the imbalanced classification datasets. Empirical results demonstrate the superiority of the proposed method that achieves higher performance on these datasets.
    \item The proposed framework is also used for the graph representation learning. Accordingly, the results of the proposed method on the link prediction and multi-label node classification problems is obtained and compared with the state-of-the-art methods like \cite{Wang2018}. These results show the superiority of the proposed method for graph representation learning.
\end{enumerate}

\section{Related Work} \label{related}
In the last few years, Generative Adversarial Networks (GANs) \cite{GAN} have received much attention in generating realistic-looking samples (usually images). The main idea of these networks is to train a generative network adversarially against a discriminator network that wants to distinguish real and fake samples.
Despite to the existing methods that employ GANs to generate samples, we intend to use the adversarial training idea to improve the classifier performance especially when confronting imbalanced datasets. Moreover, as opposed to \cite{BAGAN}, we do not create any new sample for data augmentation. Indeed, \cite{BAGAN} introduces a method to generate samples for the minority class in image classification tasks while our method seeks a classifier without generating any new sample and thus works for all types of data even including discrete features (whereas generating discrete samples for methods like \cite{BAGAN} that are based on sample generation is challenging). Although \cite{Yang} also introduces an adversarial approach in a specific domain to discriminate similar pairs of (question, answer) in the community question answering task, it uses an RL approach to learn generator while it is not necessary at all. Indeed, weights of samples are continuous that can be learned directly and we do not need to learn them via RL and sampling from the majority class examples. In the proposed method, both the generator and the discriminator are fully differentiable and thus gradient-based methods can be directly performed to optimize the objective function. Moreover, since \cite{Yang} employs exactly the objective function of GAN (for the classification problem) that may lead to highly up-weighting few negative samples (that are similar to the positive samples) and making the weights of other samples negligible (or even zero). In fact, some negative samples that may be outlier appearing among positive samples can take the leading role that is not proper. 

In this work, we propose an adversarial classifier that optimizes two criteria (instead of only a classifier or a discriminator) to solve the classification problem. A function that gives weights to negative samples is learned by the generator network and meanwhile the discriminator is trained by the positive samples and the weighted negative samples (using the weights obtained by the generator).



\section{Proposed Method} \label{proposed}
In this section, we introduce the proposed adversary-based classifier that is useful especially for imbalanced data. It is helpful in many real-world tasks like distinguishing similar pairs from dissimilar ones (when the number of dissimilar pairs is over-comming).

\subsection{Preliminaries}
The goal of Generative Adversarial Networks (GANs) \cite{GAN} is to find a generative model that produces samples from the distribution of real data. It employs the generator network to create samples and in-order to train this network, the discriminator as a critic is used to evaluate the generator network. The objective function of GANs is defined as follows \cite{GAN}:
\begin{align} \label{eq:gan_obj}
\min_{G}\max_{D}\mathbb{E}_{x \sim p_{data}(x)}[\log{D(x)}]+\\
\mathbb{E}_{x' \sim p_g(x)}[\log{(1-D(x'))}], \nonumber
\end{align}
where $p_{g}(x)$ shows the distribution of samples generated by the generator and $p_{data}(x)$ is the real data distribution. $D(x)$ represents the probability that $x$ is classified as a real sample and $D$ is trained to maximize the probability of assigning the correct label to both real examples and the generated ones by $G$. $G$ is simultaneously trained to minimize $\log{(1-D(x'))}$ on generated samples. Therefore, the generator networks attempts to generate samples that seem real and the discriminator tries to distinguish real and fake (generated) samples. 

\subsection{Objective Function} \label{objective}
Consider a two-class classification problem in which samples of the negative class are abundant. It seems that both selecting randomly a subset of negative samples and a uniform contribution of all negative samples (containing many less informative samples) may not be sufficient to reach an accurate classifier. In fact, some negative samples are more informative and make the resulted classifier more powerful.
Let $G(x)$ show the probability (or weight) assigned to the sample $x$ and $D(x)$ show the probability of classifying $x$ as positive. Thus, we propose the following objective function:
\begin{align} \label{eq:obj}
\min_{G}\max_{D}\frac{1}{|S^+|}\Sigma_{x \in S^+}\log{D(x)}+\\
\Sigma_{x \in S^-}G(x)\times \log{(1-D(x))} + \nonumber\\
\lambda \times \Sigma_{x \in S^-}G(x) \times \log(G(x)), \nonumber
\end{align}
where $S^+$ shows the positively labeled training data and $S^-$ denotes the negative ones.
In this objective function, the discriminator $D$ attempts to maximize the average log probability of assigning the correct labels to (both the positive and the negative) samples in which the weights of samples are also taken into account. The positive samples are equally weighted while the weights of negative samples are specified by the generator. On the other hand, the generator $G$ is optimized to maximize the probability of misleading the discriminator $D$. The last term in the objective function is responsible to prevent the generator to portion the weights only on the most similar samples to the positive ones. This term as the negative entropy of the generator impedes the distribution on the negative samples from converging to the distribution of positive samples. It is worth mentioning that our method does not generate any sample and $G$ only gives weight to samples.

\subsection{Training the proposed adversarial classifier}
\label{training}
To train the proposed network, $G$ and $D$ are alternatively optimized to minimize and maximize the objective function in Eq.~ \ref{eq:obj} respectively. The details of training is shown in Algorithm \ref{alg:main}. In this algorithm, we first pre-train the discriminator by minimizing cross-entropy loss function (in the first loop) and then the adversarial training is done (in the second loop). At the end of training for a mini-batch, $G$ over the samples of that batch is normalized to make sure that the output of the generator is a probability function.

\begin{algorithm}[t]
  {\small
  \SetKwData{Left}{left}\SetKwData{This}{this}\SetKwData{Up}{up}
\SetKwInOut{Input}{input}\SetKwInOut{Output}{output}
  \Input{The set of positive samples $S^+$ and negative samples $S^-$}
  \Output{$D$ (Classifier or discriminator)}

  \BlankLine
  Initialize $\theta_g$ and $\theta_d$ as the parameters of the generator and the discriminator respectively.
  
  \For{number of pre-training iterations}{
    Sample mini-batch of $m$ positive samples $\lbrace x^{+{(1)}},\ldots,x^{+{(m)}} \rbrace$ uniformly from $S^+$ \;
    Sample mini-batch of $m$ negative samples $\lbrace x^{-{(1)}},\ldots,x^{-{(m)}} \rbrace$ uniformly from $S^-$ \;
    Update the discriminator by stochastic gradient ascent:
    $\theta_d \leftarrow\theta_d+\eta \nabla_{\theta_d} \Big[\frac{1}{m}\sum_{i=1}^{m} \log(D(x^{+{(i)}})) + \frac{1}{m} \sum_{i=1}^{m}\;\log(1-D(x^{-{(i)}})) \Big]$ \;
  }
  \For{number of training iterations}{
     Sample mini-batch of $m$ positive samples $\lbrace x^{+{(1)}},\ldots,x^{+{(m)}} \rbrace$ uniformly \;
     Sample mini-batch of $m$ negative samples $\lbrace x^{-{(1)}},\ldots,x^{-{(m)}} \rbrace$ uniformly \;
     Update the discriminator by stochastic gradient ascent: 
     $\theta_d \leftarrow \theta_d+ \nabla_{\theta_d} \Big[\frac{1}{m}\sum_{i=1}^{m}\;\log(D(x^{+{(i)}})) + \gamma \sum_{i=1}^{m}{G}(x^{-{(i)}})\;\log(1-D(x^{-{(i)}})) \Big]$ \;
     Update the generator by stochastic gradient descent:
     $\theta_g \leftarrow \theta_g-\nabla_{\theta_g} \Big[\sum_{i=1}^{m}{G}(x^{-{(i)}})\;\log(1-D(x^{-{(i)}})) + \newline \lambda \sum_{i=1}^{m}{G}(x^{-{(i)}})\;\log({G}(x^{-{(i)}}))\Big]$ \nonumber \;
     }
  }
   \caption{Training Procedure of Adversarial Classifier (mini-batch stochastic gradient descent)}
 \label{alg:main}
  
\end{algorithm}


  
  

\subsection{Theoretical analysis}
\label{theory}
Analogous to \cite{GAN}, we first find the optimal discriminator $D$ for any given $G$. Then, for the optimal discriminator, an equation for the optimal generator is found. \\
\textbf{Proposition 1.} For a fixed $G$, the optimal discriminator $D$ is:
\begin{align} \label{eq:D*}
D^*_G=\frac{p_+(x)}{p_+(x)+p_{-}^{g}(x)},
\end{align}\\
where $p_+$ shows the distribution of positive samples and $p_{-}^{g}$ denotes the distribution on the negative samples after re-weighting.
\\
\textbf{Theorem 1.} For the optimal discriminator, the optimal generator must satisfy the following equation:
\begin{align} \label{eq:G*}
\frac {(p_{-}^{g}(x))^{\lambda+1}}{\int (p_{-}^{g}(x))^{\lambda+1}dx}=\frac {p_{-}^{g}(x)+p_+(x)} {2}.
\end{align}\\
The details of the proofs are available in Appendix.

First, assume that we set $\lambda=0$. In this case, we show that $G$ defines a distribution on negative samples that tends to be equal to $p_+$, i.e. the distribution of positive samples. Here, $p_+$ is considered fixed but $G$ that intends to find a new distribution on negative samples (by re-weighting them) and $D$ that is used to discriminate positive samples are trained. Analogous to \cite{GAN}, if $G$ and $D$ have enough capacity, $D$ is trained to discriminate samples of positive class and converges to Eq. \ref{eq:D*}. Moreover, $G(x)$ converges to reach $p_+(x)=p_{-}^{g}(x)$ at which
both of them cannot be improved. However, when $\lambda$ is not necessarily zero, we reach Eq. \ref{eq:G*}.

In the GAN framework, the goal is to minimize Jensen-Shannon divergence between the real data distribution and the generator distribution \cite{GAN}. Similarly, in the ARIC method, the Jensen-Shannon divergence between the distribution of positive samples and the generator distribution (that weights the existing negative samples) is minimized. 
Since the generator only gives weights to negative samples, it assigns higher weights to negative samples which are closer to the boundary and thus provides a distribution with higher masses on the boundary region.
It is important to note that the entropy term prevents the generator from learning a sharp distribution.
Moreover, in a Nash equilibrium, the discriminator learns the boundary between positive samples and weighted negative samples (in which the closer negative samples to the boundary get higher weights). Thus, it can be concluded that ARIC reduces the effect of the non-informative negative samples in classifier training.

\section{Graph Representation Learning by Adversarial Classifier} \label{graph}
In this section, we want to use the proposed avdersarial classifier for graph representation learning. First, graph representation learning and the most related methods are introduced. Then, the proposed adversarial classifier is used for graph link prediction and node classification.
\subsection{Graph Representation Learning}
\label{graph_base}

In representation learning of graph nodes, given graph structure (relations between nodes) and node features (if exist), a low dimensional latent space is found in which nodes are embedded \cite{hamilton2017representation}.
Graph node representation methods can be categorized into 3 categories \cite{cai2018comprehensive}. In the first category, methods use matrix factorization to factorize the similarity matrix of the graph (adjacency or Laplacian matrix) and extract embedding for nodes in the latent space \cite{belkin2002laplacian,kruskal1964}. Methods in the second category use optimization methods to find embedding in a way that reconstructed edges from nodes embedding are similar to the input graph edges \cite{tang2015line,tang2015pte,man2016predict}. In the third category, neural networks or deep methods are utilized to find nodes embedding via one of the following two approaches. 
In the first approach, known models like CNN, RNN, Auto-Encoder, and GAN are modified to be specialized for graphs which are non-Euclidean data \cite{NIPS2016_6081,kipf2016semi,Ying:GCN,kipf2016variational,Wang2018}. In the second approach, random walks on graph are generated (i.e. sequences of nodes are obtained from the graph structure) and then the neural networks for learning representation from the context are applied on sequences of nodes to find nodes embedding \cite{perozzi2014deepwalk,grover2016node2vec,chamberlain2017neural,chen2017harp}. 

GraphGAN \cite{Wang2018} is the most similar method to our method for graph representation learning. The purpose of GraphGAN is to find embedding for graph nodes. It includes two parts: 1) The discriminator which determines whether an edge exists between two nodes or not. It uses a sigmoid function on the dot product of the corresponding nodes embeddings for this purpose. The node embedding is updated in each iteration until convergence happens.  2) Generator which determines a connectivity distribution for each node over all other nodes of the graph. For this purpose, a graph softmax is proposed in \cite{Wang2018} to incorporate the graph structure and decrease the time complexity. 

\subsection{Adversarial Graph Representation Learning}
\label{graph_method}
In this section, the above proposed framework is used for graph representation learning. Similar to the existing approaches for graph representation learning, we also learn nodes representation from which the graph links can be predicted. Let $G=(V,E)$ be a given graph where $V$ denotes the set of nodes and $E$ shows the set of edges. The number of edges in the graphs is usually much smaller than the total number of distinct pairs of nodes, i.e. $\binom{N}{2}$

The proposed method utilizes the connected nodes in the graph as positive samples (pairs) and the disconnected ones as negative samples (pairs) but gives weight to the disconnected nodes to distinguish informative disconnected pairs from uninformative ones (for representation learning). The weights of disconnected pairs are trained adversarially such that the discrimination of negative pairs from positive ones (by the discriminator) becomes more challenging. Therefore, more informative disconnected pairs get higher weights and finally a more powerful discriminator is obtained since during training the generator makes the problem more complicated for the discriminator and thus boosts its capability. 

 \begin{figure}[!t]
 \begin{center}
 \includegraphics[width=1\columnwidth]{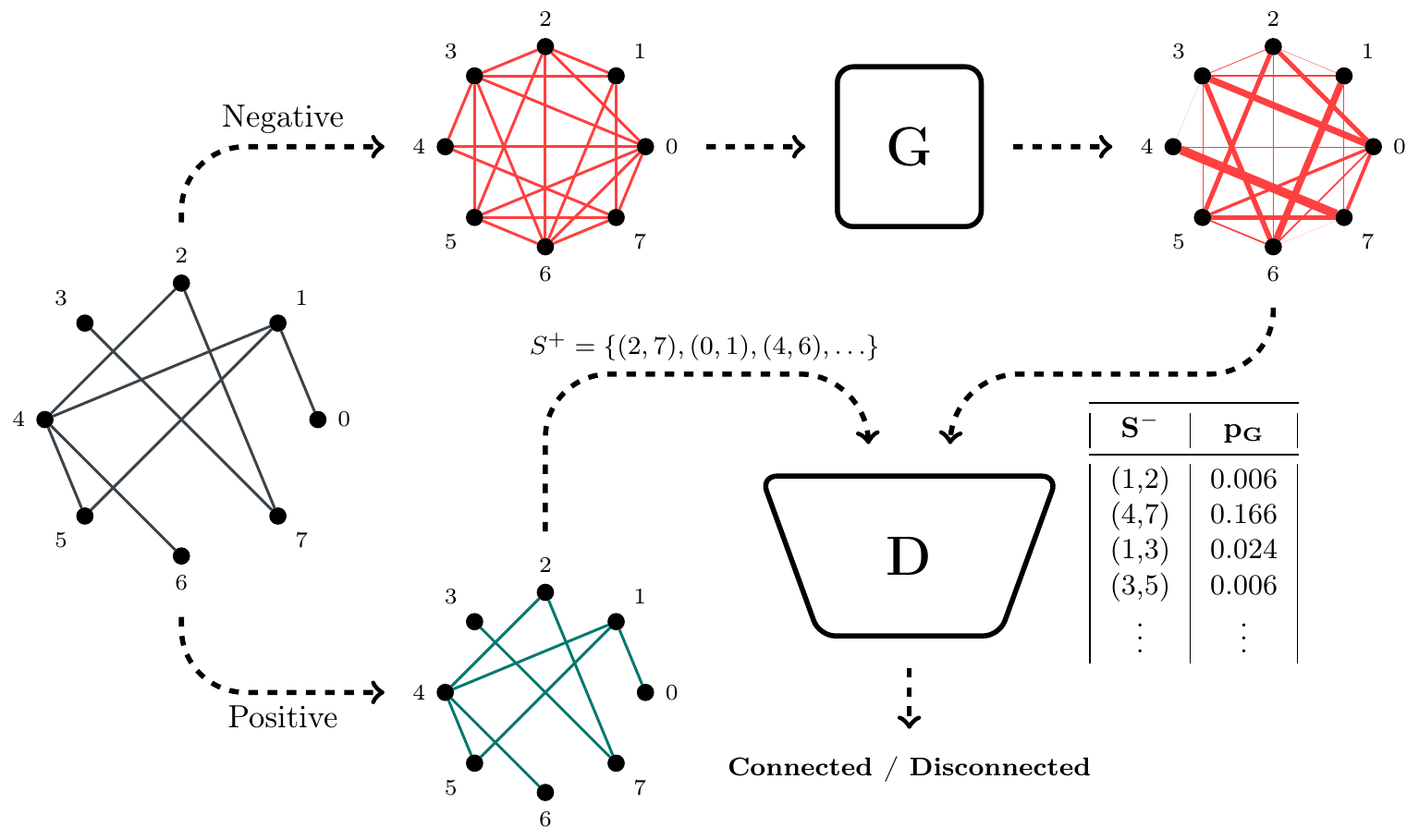}
 \caption{The proposed method for link prediction.}
 \label{fig:graph}
 \end{center}
 \end{figure}

In this problem, a function $G(v_1,v_2)$ is learned to show the effectiveness of exposing the negative pair $(v_1,v_2)$ to the discriminator as the training sample. On the other hand, the discriminator $D(v_1,v_2)$ specifies the probability of considering this pair of nodes as similar (or connected). The proposed method for the link prediction task is illustrated in Fig. \ref{fig:graph}. The main differences of the proposed method and GraphGAN \cite{Wang2018} are as follows:

\begin{enumerate}
     \item
     In the proposed method, we learn to give weights to negative samples. We do not need to use the policy gradient since the weights found by the generator are continuous and directly used in the algorithm (as weights assigned to the disconnected pairs) without need to sampling. However, in GraphGAN, the generator network is used to generate the links that seem real and for training of the network, it utilizes the policy gradient since the sampling of nodes is a discrete operator in this method. 
    \item 
    The proposed method, as opposed to GraphGAN that seeks a generator to find the distribution of connecting nodes to a specific node $v_c$, finally uses the discriminator to distinguish connected pairs from disconnected ones. In fact GraphGAN uses the trained embedding output of the generator while we use the embedding obtained by the discriminator (learned on re-weighted samples).
\end{enumerate}

The ARIC algorithm for the link prediction problem is the same as Algorithm \ref{alg:main}. The only difference is the step of training samples preparation for graphs. In graphs, a pair of connected nodes by a link represents a positive sample and a disconnected pair of nodes represents a negative sample. Thus, to generate a mini-batch from graph, we need to sample from connected and disconnected nodes of graph. On the other hand, a combiner function is needed to combine embedding of the corresponding pair of samples to provide a new feature vector for that pair. It's worth mentioning that the embedding layer of the discriminator and the generator are both initialized randomly and are trained during the training phase. The inputs to both the discriminator and the generator are the combined embedding of a pair of nodes.

\section{Experiments}
In this section, the performance of the proposed model on imbalanced classification datasets and also graph representation learning task is examined. 
\subsection{Experiments on Imbalanced Classification}
We first introduce datasets and the experimental setup for imbalanced data classification. Then, the obtained results are presented.
\subsubsection{Datasets}
The imbalanced datasets used in our experiments are presented in \cite{ding2011diversified} and provided by the imbalanced-learn python package \cite{lemaitre2017imbalanced}. Pen\_digit, letter\_img, webpage, mammography, and protetin\_homo are the five largest datasets that are chosen to evaluate the  ability of the proposed method on them. These datatsets also have different amount of imbalanced-ration (IR). The properties of these datasets are provided in Table \ref{tab:datasets}.
We use $60\%$ of the dataset for training, $20\%$ for test, and $20\%$ for validation.

\begin{table}[!t]
\caption{The properties of binary-class imbalanced datasets: the imbalanced ration (IR), the whole number of samples, and the number of attributes.}
\label{tab:datasets}
\centering
\small
\begin{tabular}{ |c|c|c|c| }
\hline
Dataset Name & IR & \#~Samples & \#~Attributes\\ \hline
pen\_digits & 9.4:1 & 10992 & 16  \\ \hline
letter\_img & 26:1 & 20000 & 16 \\ \hline
webpage & 33:1 & 34780 & 300  \\ \hline
mammography & 42:1 & 11183 & 6  \\ \hline
protein\_homo & 111.5:1 & 145751 & 74  \\ \hline
\end{tabular}
\end{table}

\subsubsection{Experimental Setup}

Evaluation metrics are accuracy, AUC-ROC, F1-score, and precision to evaluate the classifiers on imbalanced datasets. From the imbalanced-learn python package \cite{lemaitre2017imbalanced}, we use Instance Hardness Threshold \cite{smith2014instance}, SVM SMOTE \cite{nguyen2011borderline}, SMOTE \cite{chawla2002smote}, and ADADYN \cite{he2008adasyn} as powerful imbalanced classification methods. All the methods use re-sampling technique to solve the imbalance problem in classification. SMOTE, SVM SMOTE and ADASYN are based on over-sampling and Instance Hardness Threshold is based on the under-sampling approach. 

In all the methods (including the proposed method and its competitors), logistic regression is used as the discriminator network and in the proposed method one of the following network architectures is used for the generator based on the results on the validation set. The first architecture contains 3-layer network with 64, 32 and 32 nodes in its layers, and the second one is a 6-layer network with 10, 8, 8, 6, 6, and 6 nodes in each layer respectively.
In the current experiences, the deeper structure with 6 layers has better performance on pen\_digits and protein\_homo datasets and the shallower one is selected for the other datasets (according to the results on the validation set).


\subsubsection{Results}
The obtained results on imbalanced classification benchmarks are summarized in Table \ref{tab:benchmark_reults2}. We show that our ARIC method generally outperforms the other methods on these datasets. 

\begin{table*}[!t]
\caption{Performance of the imbalanced classification methods that use Logistic Regression as the discriminator on different datasets.}
\label{tab:benchmark_reults2}
\centering
\begin{tabular}{llllll}
\hline
\multicolumn{1}{|l|}{Dataset}                                 & \multicolumn{1}{l|}{Model}                 & \multicolumn{1}{l|}{Accuracy}               & \multicolumn{1}{l|}{AUC}                   & \multicolumn{1}{l|}{Precision}             & \multicolumn{1}{l|}{F1-Score}              \\ 
\noalign{\global\arrayrulewidth=0.3mm}
\hline 
\noalign{\global\arrayrulewidth=0.1mm}
\multicolumn{1}{|l|}{\multirow{6}{*}{pen\_digits}} & \multicolumn{1}{l|}{ Instance Hardness Threshold} & \multicolumn{1}{l|}{0.9604} &  \multicolumn{1}{l|}{0.8894} & \multicolumn{1}{l|}{0.7830} & \multicolumn{1}{l|}{0.7924} \\ \cline{2-6} 
\multicolumn{1}{|l|}{}                  & \multicolumn{1}{l|}{SVM SMOTE} & \multicolumn{1}{l|}{0.9026} & \multicolumn{1}{l|}{0.9289} & \multicolumn{1}{l|}{0.4914} & \multicolumn{1}{l|}{0.6503} \\ \cline{2-6} 
\multicolumn{1}{|l|}{}                  & \multicolumn{1}{l|}{SMOTE} & \multicolumn{1}{l|}{0.9359} &  \multicolumn{1}{l|}{0.9235} & \multicolumn{1}{l|}{0.6065} & \multicolumn{1}{l|}{0.7273} \\ \cline{2-6} 
\multicolumn{1}{|l|}{}                  & \multicolumn{1}{l|}{ADASYN} & \multicolumn{1}{l|}{0.8799} &  \multicolumn{1}{l|}{0.8580} & \multicolumn{1}{l|}{0.4289} & \multicolumn{1}{l|}{0.5658} \\ \cline{2-6} 
\multicolumn{1}{|l|}{}                  & \multicolumn{1}{l|}{Imbalance Classifier} & \multicolumn{1}{l|}{\textbf{0.9636}} &  \multicolumn{1}{l|}{0.8652} & \multicolumn{1}{l|}{\textbf{0.8508}} & \multicolumn{1}{l|}{0.7938} \\ \cline{2-6} 

\multicolumn{1}{|l|}{}                  & \multicolumn{1}{l|}{ARIC} & \multicolumn{1}{l|}{\textbf{0.9636}} &  \multicolumn{1}{l|}{\textbf{0.9722}} & \multicolumn{1}{l|}{0.7981} & \multicolumn{1}{l|}{\textbf{0.8095}} \\ 
\noalign{\global\arrayrulewidth=0.3mm}
\hline 
\noalign{\global\arrayrulewidth=0.1mm}
\multicolumn{1}{|l|}{\multirow{6}{*}{letter\_img}} & \multicolumn{1}{l|}{ Instance Hardness Threshold} & \multicolumn{1}{l|}{0.9825} &  \multicolumn{1}{l|}{0.8176} & \multicolumn{1}{l|}{0.8165} & \multicolumn{1}{l|}{\textbf{0.7177}} \\ \cline{2-6} 
\multicolumn{1}{|l|}{}                  & \multicolumn{1}{l|}{SVM SMOTE} & \multicolumn{1}{l|}{0.9237} & \multicolumn{1}{l|}{0.9536} & \multicolumn{1}{l|}{0.3114} & \multicolumn{1}{l|}{0.4732} \\ \cline{2-6} 
\multicolumn{1}{|l|}{}                  & \multicolumn{1}{l|}{SMOTE} & \multicolumn{1}{l|}{0.9483} &  \multicolumn{1}{l|}{0.9559} & \multicolumn{1}{l|}{0.3988} & \multicolumn{1}{l|}{0.5642} \\ \cline{2-6} 
\multicolumn{1}{|l|}{}                  & \multicolumn{1}{l|}{ADASYN} & \multicolumn{1}{l|}{0.9265} &  \multicolumn{1}{l|}{0.9377} & \multicolumn{1}{l|}{0.3150} & \multicolumn{1}{l|}{0.4731} \\ \cline{2-6} 
\multicolumn{1}{|l|}{}                  & \multicolumn{1}{l|}{Imbalance Classifier} & \multicolumn{1}{l|}{0.9818} &  \multicolumn{1}{l|}{0.7894} & \multicolumn{1}{l|}{0.8438} & \multicolumn{1}{l|}{0.6894} \\ \cline{2-6} 
\multicolumn{1}{|l|}{}                  & \multicolumn{1}{l|}{ARIC} & \multicolumn{1}{l|}{\textbf{0.9835}} &  \multicolumn{1}{l|}{\textbf{0.9751}} & \multicolumn{1}{l|}{\textbf{0.8925}} & \multicolumn{1}{l|}{0.7155} \\ 
\noalign{\global\arrayrulewidth=0.3mm}
\hline 
\noalign{\global\arrayrulewidth=0.1mm} 
\multicolumn{1}{|l|}{\multirow{6}{*}{webpage}} & \multicolumn{1}{l|}{ Instance Hardness Threshold} & \multicolumn{1}{l|}{0.9858} &  \multicolumn{1}{l|}{0.8676} & \multicolumn{1}{l|}{0.7151} & \multicolumn{1}{l|}{0.7288} \\ \cline{2-6} 
\multicolumn{1}{|l|}{}                  & \multicolumn{1}{l|}{SVM SMOTE} & \multicolumn{1}{l|}{0.9779} &  \multicolumn{1}{l|}{0.9315} & \multicolumn{1}{l|}{0.5430} & \multicolumn{1}{l|}{0.6723} \\ \cline{2-6} 
\multicolumn{1}{|l|}{}                  & \multicolumn{1}{l|}{SMOTE} & \multicolumn{1}{l|}{0.9612} &  \multicolumn{1}{l|}{0.9338} & \multicolumn{1}{l|}{0.3904} & \multicolumn{1}{l|}{0.5455} \\ \cline{2-6} 
\multicolumn{1}{|l|}{}                  & \multicolumn{1}{l|}{ADASYN} & \multicolumn{1}{l|}{0.9470} &  \multicolumn{1}{l|}{0.9238} & \multicolumn{1}{l|}{0.3145} & \multicolumn{1}{l|}{0.4660} \\ \cline{2-6} 
\multicolumn{1}{|l|}{}                  & \multicolumn{1}{l|}{Imbalance Classifier} & \multicolumn{1}{l|}{0.9876} &  \multicolumn{1}{l|}{0.8006} & \multicolumn{1}{l|}{0.8780} & \multicolumn{1}{l|}{0.7152} \\ \cline{2-6} 
\multicolumn{1}{|l|}{}                  & \multicolumn{1}{l|}{ARIC} & \multicolumn{1}{l|}{\textbf{0.9894}} &  \multicolumn{1}{l|}{\textbf{0.9734}} & \multicolumn{1}{l|}{\textbf{0.8889}} & \multicolumn{1}{l|}{\textbf{0.7643}} \\
\noalign{\global\arrayrulewidth=0.3mm}
\hline 
\noalign{\global\arrayrulewidth=0.1mm}
\multicolumn{1}{|l|}{\multirow{6}{*}{mammography}} & \multicolumn{1}{l|}{ Instance Hardness Threshold} & \multicolumn{1}{l|}{0.9830} & \multicolumn{1}{l|}{0.7220} & \multicolumn{1}{l|}{0.6667} & \multicolumn{1}{l|}{\textbf{0.5366}} \\ \cline{2-6} 
\multicolumn{1}{|l|}{}                  & \multicolumn{1}{l|}{SVM SMOTE} & \multicolumn{1}{l|}{0.9334} &  \multicolumn{1}{l|}{0.8961} & \multicolumn{1}{l|}{0.2283} & \multicolumn{1}{l|}{0.3605} \\ \cline{2-6} 
\multicolumn{1}{|l|}{}                  & \multicolumn{1}{l|}{SMOTE} & \multicolumn{1}{l|}{0.8820} &  \multicolumn{1}{l|}{0.8898} & \multicolumn{1}{l|}{0.1452} & \multicolumn{1}{l|}{0.2500} \\ \cline{2-6} 
\multicolumn{1}{|l|}{}                  & \multicolumn{1}{l|}{ADASYN} & \multicolumn{1}{l|}{0.8234} &  \multicolumn{1}{l|}{0.8898} & \multicolumn{1}{l|}{0.1068} & \multicolumn{1}{l|}{0.1922} \\ \cline{2-6} 
\multicolumn{1}{|l|}{}                  & \multicolumn{1}{l|}{Imbalance Classifier} & \multicolumn{1}{l|}{\textbf{0.9835}} &  \multicolumn{1}{l|}{0.6624} & \multicolumn{1}{l|}{\textbf{0.8000}} & \multicolumn{1}{l|}{0.4638} \\ \cline{2-6} 
\multicolumn{1}{|l|}{}                  & \multicolumn{1}{l|}{ARIC} & \multicolumn{1}{l|}{0.9830} &  \multicolumn{1}{l|}{\textbf{0.9360}} & \multicolumn{1}{l|}{0.6667} & \multicolumn{1}{l|}{\textbf{0.5366}} \\ 
\noalign{\global\arrayrulewidth=0.3mm}
\hline 
\noalign{\global\arrayrulewidth=0.1mm}
\multicolumn{1}{|l|}{\multirow{6}{*}{protein\_homo}} & \multicolumn{1}{l|}{ Instance Hardness Threshold} & \multicolumn{1}{l|}{\textbf{0.9974}} &  \multicolumn{1}{l|}{0.8985} & \multicolumn{1}{l|}{0.9048} & \multicolumn{1}{l|}{\textbf{0.8479}} \\ \cline{2-6} 
\multicolumn{1}{|l|}{}                  & \multicolumn{1}{l|}{SVM SMOTE} & \multicolumn{1}{l|}{0.9915}  & \multicolumn{1}{l|}{0.9258} & \multicolumn{1}{l|}{0.5172} & \multicolumn{1}{l|}{0.6456} \\ \cline{2-6} 
\multicolumn{1}{|l|}{}                  & \multicolumn{1}{l|}{SMOTE} & \multicolumn{1}{l|}{0.9726} &  \multicolumn{1}{l|}{0.9484} & \multicolumn{1}{l|}{0.2373} & \multicolumn{1}{l|}{0.3775} \\ \cline{2-6} 
\multicolumn{1}{|l|}{}                  & \multicolumn{1}{l|}{ADASYN} & \multicolumn{1}{l|}{0.9642} &  \multicolumn{1}{l|}{0.9498} & \multicolumn{1}{l|}{0.1925} & \multicolumn{1}{l|}{0.3192} \\ \cline{2-6} 
\multicolumn{1}{|l|}{}                  & \multicolumn{1}{l|}{Imbalance Classifier} & \multicolumn{1}{l|}{0.9972} &  \multicolumn{1}{l|}{0.8643} & \multicolumn{1}{l|}{\textbf{0.9409}} & \multicolumn{1}{l|}{0.8215} \\ \cline{2-6} 
\multicolumn{1}{|l|}{}                  & \multicolumn{1}{l|}{ARIC} & \multicolumn{1}{l|}{0.9969	} &  \multicolumn{1}{l|}{\textbf{0.9801}} & \multicolumn{1}{l|}{0.9212} & \multicolumn{1}{l|}{0.8043} \\ \hline

\end{tabular}
\end{table*}


\subsection{Experiments on Graph Embedding} \label{graph_experiments}
In this section, we conduct experiments to evaluate performance of trained nodes embedding by the proposed method on Link prediction and multi-label node classification that are two standard supervised learning tasks on graphs. 
We compare our method with the state-of-the-art methods that learn embedding for graph nodes (on link prediction and node classification tasks). We use the proposed method to learn a capable classifier (or discriminator) that can distinguish connected pairs of nodes from disconnected ones. The embedding layer of the discriminator can be used to embed nodes of the graph appropriately.

\subsubsection{Graph Datasets} \label{graph_data}
Similar to GraphGAN \cite{Wang2018}, the ability of methods for link prediction is tested on the following datasets:
\begin{itemize}
    \item arXiv-AstroPh \cite{leskovec2015snap}: The network is generated from papers submitted to the Astro Physics category of e-print arXiv. Nodes are scientists and edges show the collaboration between two scientists in a paper. The network contains $18772$ nodes and $198110$ edges.
    \item arXiv-GrQc \cite{leskovec2015snap}: This network is about papers submmited to General Relativity and Quantum Cosmology categories of arXiv. Same as the previous network, nodes present scientists and edges show collaboration. The network contains $5242$ nodes and $14496$ edges.
\end{itemize}

Moreover, as in \cite{Wang2018}, the following datasets are utilized to test the ability of the proposed method on multi-label node classification task:
\begin{itemize}
    \item BlogCatalog \cite{zafarani2009social}: The network is constructed based on the social relationship between bloggers on the BlogCatalog website. The labels of bloggers shows the blogger interests and extracted from meta-data provided by the bloggers. The network contains $10312$ nodes and $333983$ edges and $39$ different labels. 
    \item Wikipedia \cite{mahoney2011large}: This network is based on the co-occurrence of words appearing in the first $10^9$ bytes of the English Wikipedia dump. Labels of the network show the Part-of-Speech (POS) tags of words. The network contains $4777$ nodes and $184812$ edges with $40$ different labels.
\end{itemize}

\subsubsection{Experimental Setup}
To evaluate the performance of the learned features by our method, the results of the proposed method is compared with those of the following feature learning algorithms:
\begin{itemize}
    \item DeepWalk\footnote{https://github.com/phanein/deepwalk} \cite{perozzi2014deepwalk}: This method is based on the uniform random walk on the graph and uses Skip-Gram to learn $d$-dimensional feature representation. 
    \item Line\footnote{https://github.com/tangjianpku/LINE} \cite{tang2015line}: The goal of this method is to preserve the first and the second order proximities. For this purpose, $\frac{d}{2}$ features from $d$ features are learned by immediate neighbors of a node and the other $\frac{d}{2}$ of features are learned by 2-hop neighbors.
    \item node2vec\footnote{https://github.com/aditya-grover/node2vec} \cite{grover2016node2vec}: It is the generalized version of DeepWalk and uses a biased random walk instead of uniform.
    \item GraphGAN\footnote{https://github.com/hwwang55/GraphGAN} \cite{Wang2018}: As we said before, the method is a GAN-based approach. It learns the generator which determines a connectivity distribution for each node over all other nodes of the graph to find nodes embedding. 
\end{itemize}
The setting of the experiments are chosen as in GraphGAN \cite{Wang2018}. Parameters of all baseline methods are set to their default values. The output embedding of the last round of GraphGAN is used as the final embedding. In all benchmarks, optimization is done using stochastic gradient descent to update the parameters. In ARIC, we set $1e-3$ as the learning rate for the discriminator, $1e-5$ as the learning rate for the generator, and $1024$ as the batch size in each iteration. In Algorithm \ref{alg:main}, $\gamma$ (the negative class weight) and $\lambda$ (the entropy weight) are set to $1e-3$ and $1e-1$ respectively. All the parameters of the proposed method are chosen by cross-validation. The combiner function (that is used to combine embedding for a pair of nodes) is chosen to be the dot product for the discriminator and the concatenation for the generator.  The dimension of embedding for all methods is set to 20.\\
In the proposed method, a network with 2 layers is selected for discriminator where the first layer is used for learning input embeddings (the input is a one-hot vector) and the second layer contains only one node for classification. On the other hand, a 3-layer network, with 64, 32 and 32 nodes in its layers, is used as generator. Discriminator is selected as simple as possible to keep the simplicity of the model.


\textbf{Link Prediction.}
In the link prediction problem, a network is given and the goal is to use nodes embedding and predict whether an edge exists between two nodes or not. We chose similar settings to GraphGAN \cite{Wang2018}. $10\%$ of the original graph edges are selected randomly for test and the rest of the network is utilized for training. To generate negative samples for test, those pair of nodes that are not connected in the original graph are chosen. The number of positive and negative samples in the test set are equal. To compare our discriminator results with other methods, after learning embedding for all nodes, a logistic regression model is trained for each of the other methods to predict the edge existence between the node pairs in the test set (as in \cite{Wang2018}). We measure the quality of our work by Accuracy and Macro-F1 score as the most commonly used measures for the link prediction. Similar to \cite{Wang2018}), in other methods, median is used as a threshold to map the output of logistic regression discriminator to (zero and one) labels. Since the number of positive and negative edges in the test set are equal, it's the best threshold and the reported results of the other methods using this threshold are the best obtained results for them. In our method, we simply round the probability obtained by the discriminator to prevent the bias toward the size of the test samples in each class.\\

\textbf{Multi-label Node Classification.}
In multi-label node classification problem, there are $C$ classes and each node belongs to one or more classes. For evaluation, we hide $p$ percent of node labels and train a logistic-regression model by one vs. all strategy on learned features of nodes. The goal is to predict the labels of hidden nodes. 
The performance of all the methods are compared when the percent of train samples is set to $90\%$. Since the application is multi-label classification, Micro-F1 and Macro-F1 scores are chosen as the most commonly used measures for this purpose.

\subsubsection{Results} 
The results of the link prediction on arXiv-AstroPh and arXiv-GrQc are summarized in Table \ref{tab:lp}. These results indicate that ARIC outperforms other methods significantly and achieves improvements on both the datasets. In fact, in the link prediction problem, we intend to find a new latent space in which connected pairs of nodes can be discriminated from disconnected ones. In the other words, we want to embed nodes such that similar pairs of nodes and dissimilar ones can be distinguished accurately. If we consider similar pairs as positive and dissimilar pairs as negative samples, the resulted classification problem is usually extremely imbalanced. Moreover, some dissimilar pairs are more informative than others for discriminating similar from dissimilar. From the obtained results, we can conclude that learning to give weights to the negative samples is an efficient way to find a better embedding and thus more accurate discriminator in the new latent space.
\begin{table}[!t]
\caption{Accuracy and Macro-F1 scores of embedding methods on arXiv-AstroPh and arXiv-GrQc for the link prediction.}
\label{tab:lp}
\begin{tabular}{ |c|c|c|c|c| }
\hline
  & \multicolumn{2}{|c|}{arXiv-AstroPh}  & \multicolumn{2}{|c|}{arXiv-GrQc} \\ \hline
Method & Acc & Macro-F1 & Acc & Macro-F1 \\ \hline
Deep Walk & 0.8252 & 0.8202 & 0.8306 & 0.8257 \\ \hline
Line & 0.7080 & 0.6997 & 0.7330 & 0.7254 \\ \hline
node2vec & 0.8249 & 0.8199 & 0.8297 & 0.8248 \\ \hline
GraphGAN & 0.8186 & 0.8134 & 0.8324 & 0.8276 \\ \hline
ARIC & \textbf{0.9190} & \textbf{0.9126} & \textbf{0.8978} & \textbf{0.9177} \\ \hline
\end{tabular}
\end{table}

We run the methods for 10 different shuffles of node labels. The mean and standard deviation of the results of multi-label node classification on BlogCatalog and Wikipedia datasets are shown in Table. \ref{tab:nc}. The node classification application is based on the embedding problem, because the labels of nodes is related to the structure of the input graph. Therefore, the results of the node classification application is influenced by the ability of node embedding in discriminating similar (connected) and dissimilar (disconnected) nodes. If we find a better embedding for nodes, it's expected to have better results of multi-label node classification. According to Table. \ref{tab:nc}, ARIC has shown better results than the other methods and it can be concluded that the learned embedding by the ARIC shows the capability of modeling the structure of the graph in the latent space better than its competitors.   
\begin{table}[!t]
\caption{Micro-F1 and Macro-F1 scores of embedding methods on BlogCatalog and Wikipedia datasets for the node classification.}
\label{tab:nc}
\small
\begin{tabular}{|>{\centering\arraybackslash}m{1.3cm}|>{\centering\arraybackslash}m{1.3cm}|>{\centering\arraybackslash}m{1.3cm}|>{\centering\arraybackslash}m{1.3cm}|>{\centering\arraybackslash}m{1.3cm}| }
\hline
  & \multicolumn{2}{|c|}{BlogCatalog}  & \multicolumn{2}{|c|}{Wikipedia} \\ \hline
\multicolumn{1}{|c|}{Method} & \multicolumn{1}{|c|}{Micro-F1} & \multicolumn{1}{|c|}{Macro-F1} &  \multicolumn{1}{|c|}{Micro-F1} &  \multicolumn{1}{|c|}{Macro-F1} \\ \hline
\multicolumn{1}{|c|}{Deep Walk} & 0.375 \newline $\pm$ 0.010 & 0.193 \newline $\pm$ 0.007 & 0.472 \newline $\pm$ 0.025  & 0.079 \newline $\pm$ 0.004 \\ \hline
\multicolumn{1}{|c|}{Line} & 0.254\newline $\pm$ 0.009 & 0.085\newline $\pm$ 0.003 & 0.436\newline $\pm$ 0.030 & 0.062\newline $\pm$ 0.007 \\ \hline
\multicolumn{1}{|c|}{node2vec} & 0.373\newline $\pm$ 0.009 & 0.194\newline $\pm$ 0.009 & 0.469\newline $\pm$ 0.026 & 0.078\newline $\pm$ 0.004\\ \hline
\multicolumn{1}{|c|}{GraphGAN} & 0.284\newline $\pm$ 0.0.008 & 0.127\newline $\pm$ 0.010 & 0.472\newline $\pm$ 0.025 & 0.079\newline $\pm$ 0.004 \\ \hline
\multicolumn{1}{|c|}{ARIC} & \textbf{0.380\newline $\pm$ 0.009} & \textbf{0.198\newline $\pm$ 0.008} & \textbf{0.481\newline $\pm$ 0.022} & \textbf{0.084\newline $\pm$ 0.005}  \\ \hline
\end{tabular}
\end{table}

\section{Conclusion} \label{conclusion}
In this paper, we proposed an adversarial learning method for imbalanced classification problems. The proposed ARIC method is able to learn the importance of incorporation of each negative sample (i.e. each sample of the majority class) in the classification loss function. The generator network give weights to the negative samples and the discriminator attempts to find a classifier using the positive and the weighted negative samples. These networks are trained in an adversarial manner and boost each other capability. Experimental results on imbalanced classification problems and also graph representation learning shows the higher performance of the proposed method compared to the other methods.\\
In the future work, we intend to learn two generator networks for the two-class classification problem where each of these generators gives weights to the samples of one the classes. Therefore, each generator specifies more informative samples for one of the classes.  
{\small
\bibliographystyle{aaai}
\bibliography{Adversarial.bib}

\begin{thebibliography}{}

\bibitem[\protect\citeauthoryear{Ando and Huang}{2017}]{deep_over}
Ando, S., and Huang, C.~Y.
\newblock 2017.
\newblock Deep over-sampling framework for classifying imbalanced data.
\newblock In {\em Joint European Conference on Machine Learning and Knowledge
  Discovery in Databases},  770--785.
\newblock Springer.

\bibitem[\protect\citeauthoryear{Belkin and Niyogi}{2002}]{belkin2002laplacian}
Belkin, M., and Niyogi, P.
\newblock 2002.
\newblock Laplacian eigenmaps and spectral techniques for embedding and
  clustering.
\newblock In {\em Advances in neural information processing systems},
  585--591.

\bibitem[\protect\citeauthoryear{Cai, Zheng, and
  Chang}{2018}]{cai2018comprehensive}
Cai, H.; Zheng, V.~W.; and Chang, K.
\newblock 2018.
\newblock A comprehensive survey of graph embedding: problems, techniques and
  applications.
\newblock {\em IEEE Transactions on Knowledge and Data Engineering}.

\bibitem[\protect\citeauthoryear{Chamberlain, Clough, and
  Deisenroth}{2017}]{chamberlain2017neural}
Chamberlain, B.~P.; Clough, J.; and Deisenroth, M.~P.
\newblock 2017.
\newblock Neural embeddings of graphs in hyperbolic space.
\newblock {\em arXiv preprint arXiv:1705.10359}.

\bibitem[\protect\citeauthoryear{Chawla \bgroup et al\mbox.\egroup
  }{2002}]{chawla2002smote}
Chawla, N.~V.; Bowyer, K.~W.; Hall, L.~O.; and Kegelmeyer, W.~P.
\newblock 2002.
\newblock Smote: synthetic minority over-sampling technique.
\newblock {\em Journal of artificial intelligence research} 16:321--357.

\bibitem[\protect\citeauthoryear{Chen \bgroup et al\mbox.\egroup
  }{2017}]{chen2017harp}
Chen, H.; Perozzi, B.; Hu, Y.; and Skiena, S.
\newblock 2017.
\newblock Harp: hierarchical representation learning for networks.
\newblock {\em arXiv preprint arXiv:1706.07845}.

\bibitem[\protect\citeauthoryear{Defferrard, Bresson, and
  Vandergheynst}{2016}]{NIPS2016_6081}
Defferrard, M.; Bresson, X.; and Vandergheynst, P.
\newblock 2016.
\newblock Convolutional neural networks on graphs with fast localized spectral
  filtering.
\newblock In {\em Advances in Neural Information Processing Systems},
  3844--3852.

\bibitem[\protect\citeauthoryear{Ding}{2011}]{ding2011diversified}
Ding, Z.
\newblock 2011.
\newblock {\em Diversified ensemble classifiers for highly imbalanced data
  learning and their application in bioinformatics}.
\newblock Ph.D. Dissertation, Georgia State University.

\bibitem[\protect\citeauthoryear{Dong, Gong, and Zhu}{2018}]{deep_imbalanced}
Dong, Q.; Gong, S.; and Zhu, X.
\newblock 2018.
\newblock Imbalanced deep learning by minority class incremental rectification.
\newblock {\em IEEE Transactions on Pattern Analysis and Machine Intelligence}.

\bibitem[\protect\citeauthoryear{Goodfellow \bgroup et al\mbox.\egroup
  }{2014}]{GAN}
Goodfellow, I.; Pouget-Abadie, J.; Mirza, M.; Xu, B.; Warde-Farley, D.; Ozair,
  S.; Courville, A.; and Bengio, Y.
\newblock 2014.
\newblock Generative adversarial nets.
\newblock In {\em Advances in neural information processing systems 27},
  2672--2680.

\bibitem[\protect\citeauthoryear{Grover and
  Leskovec}{2016}]{grover2016node2vec}
Grover, A., and Leskovec, J.
\newblock 2016.
\newblock node2vec: Scalable feature learning for networks.
\newblock In {\em Proceedings of the 22nd ACM SIGKDD international conference
  on Knowledge discovery and data mining},  855--864.
\newblock ACM.

\bibitem[\protect\citeauthoryear{Hamilton, Ying, and
  Leskovec}{2017}]{hamilton2017representation}
Hamilton, W.~L.; Ying, R.; and Leskovec, J.
\newblock 2017.
\newblock Representation learning on graphs: Methods and applications.
\newblock {\em arXiv preprint arXiv:1709.05584}.

\bibitem[\protect\citeauthoryear{He \bgroup et al\mbox.\egroup
  }{2008}]{he2008adasyn}
He, H.; Bai, Y.; Garcia, E.~A.; and Li, S.
\newblock 2008.
\newblock Adasyn: Adaptive synthetic sampling approach for imbalanced learning.
\newblock In {\em Neural Networks, 2008. IJCNN 2008.(IEEE World Congress on
  Computational Intelligence). IEEE International Joint Conference on},
  1322--1328.
\newblock IEEE.

\bibitem[\protect\citeauthoryear{Huang \bgroup et al\mbox.\egroup
  }{2016}]{deep_cvpr}
Huang, C.; Li, Y.; Change~Loy, C.; and Tang, X.
\newblock 2016.
\newblock Learning deep representation for imbalanced classification.
\newblock In {\em Proceedings of the IEEE Conference on Computer Vision and
  Pattern Recognition},  5375--5384.

\bibitem[\protect\citeauthoryear{Kipf and Welling}{2016a}]{kipf2016semi}
Kipf, T.~N., and Welling, M.
\newblock 2016a.
\newblock Semi-supervised classification with graph convolutional networks.
\newblock {\em arXiv preprint arXiv:1609.02907}.

\bibitem[\protect\citeauthoryear{Kipf and Welling}{2016b}]{kipf2016variational}
Kipf, T.~N., and Welling, M.
\newblock 2016b.
\newblock Variational graph auto-encoders.
\newblock {\em arXiv preprint arXiv:1611.07308}.

\bibitem[\protect\citeauthoryear{Krawczyk}{2016}]{Bartosz}
Krawczyk, B.
\newblock 2016.
\newblock Learning from imbalanced data: open challenges and future directions.
\newblock {\em Progress in Artificial Intelligence} 5(4):221--232.

\bibitem[\protect\citeauthoryear{Kruskal}{1964}]{kruskal1964}
Kruskal, J.~B.
\newblock 1964.
\newblock Multidimensional scaling by optimizing goodness of fit to a nonmetric
  hypothesis.
\newblock {\em Psychometrika} 29(1):1--27.

\bibitem[\protect\citeauthoryear{Lema{\^\i}tre, Nogueira, and
  Aridas}{2017}]{lemaitre2017imbalanced}
Lema{\^\i}tre, G.; Nogueira, F.; and Aridas, C.~K.
\newblock 2017.
\newblock Imbalanced-learn: A python toolbox to tackle the curse of imbalanced
  datasets in machine learning.
\newblock {\em The Journal of Machine Learning Research} 18(1):559--563.

\bibitem[\protect\citeauthoryear{Leskovec and Krevl}{2015}]{leskovec2015snap}
Leskovec, J., and Krevl, A.
\newblock 2015.
\newblock Snap datasets:stanford large network dataset collection.
\newblock {\em URL:http://snap.stanford.edu/data}.

\bibitem[\protect\citeauthoryear{Mahoney}{2011}]{mahoney2011large}
Mahoney, M.
\newblock 2011.
\newblock Large text compression benchmark.
\newblock {\em URL: http://www.mattmahoney. net/text/text.html}.

\bibitem[\protect\citeauthoryear{Man \bgroup et al\mbox.\egroup
  }{2016}]{man2016predict}
Man, T.; Shen, H.; Liu, S.; Jin, X.; and Cheng, X.
\newblock 2016.
\newblock Predict anchor links across social networks via an embedding
  approach.
\newblock In {\em IJCAI}, volume~16,  1823--1829.

\bibitem[\protect\citeauthoryear{Mariani \bgroup et al\mbox.\egroup
  }{2018}]{BAGAN}
Mariani, G.; Scheidegger, F.; Istrate, R.; Bekas, C.; and Malossi, C.
\newblock 2018.
\newblock Bagan: Data augmentation with balancing gan.
\newblock {\em arXiv preprint arXiv:1803.09655}.

\bibitem[\protect\citeauthoryear{Nguyen, Cooper, and
  Kamei}{2011}]{nguyen2011borderline}
Nguyen, H.~M.; Cooper, E.~W.; and Kamei, K.
\newblock 2011.
\newblock Borderline over-sampling for imbalanced data classification.
\newblock {\em International Journal of Knowledge Engineering and Soft Data
  Paradigms} 3(1):4--21.

\bibitem[\protect\citeauthoryear{Perozzi, Al-Rfou, and
  Skiena}{2014}]{perozzi2014deepwalk}
Perozzi, B.; Al-Rfou, R.; and Skiena, S.
\newblock 2014.
\newblock Deepwalk: Online learning of social representations.
\newblock In {\em Proceedings of the 20th ACM SIGKDD international conference
  on Knowledge discovery and data mining},  701--710.
\newblock ACM.

\bibitem[\protect\citeauthoryear{Smith, Martinez, and
  Giraud-Carrier}{2014}]{smith2014instance}
Smith, M.~R.; Martinez, T.; and Giraud-Carrier, C.
\newblock 2014.
\newblock An instance level analysis of data complexity.
\newblock {\em Machine learning} 95(2):225--256.

\bibitem[\protect\citeauthoryear{Tang \bgroup et al\mbox.\egroup
  }{2015}]{tang2015line}
Tang, J.; Qu, M.; Wang, M.; Zhang, M.; Yan, J.; and Mei, Q.
\newblock 2015.
\newblock Line: Large-scale information network embedding.
\newblock In {\em Proceedings of the 24th International Conference on World
  Wide Web},  1067--1077.
\newblock International World Wide Web Conferences Steering Committee.

\bibitem[\protect\citeauthoryear{Tang, Qu, and Mei}{2015}]{tang2015pte}
Tang, J.; Qu, M.; and Mei, Q.
\newblock 2015.
\newblock Pte: Predictive text embedding through large-scale heterogeneous text
  networks.
\newblock In {\em Proceedings of the 21th ACM SIGKDD International Conference
  on Knowledge Discovery and Data Mining},  1165--1174.
\newblock ACM.

\bibitem[\protect\citeauthoryear{Wang \bgroup et al\mbox.\egroup
  }{2018}]{Wang2018}
Wang, H.; Wang, J.; Wang, J.; Zhao, M.; Zhang, W.; Zhang, F.; Xie, X.; and Guo,
  M.
\newblock 2018.
\newblock Graphgan: Graph representation learning with generative adversarial
  nets.
\newblock In {\em Proceedings of the AAAI Conference on Artificial
  Intelligence. AAAI Conference on Artificial Intelligence}.

\bibitem[\protect\citeauthoryear{Yang \bgroup et al\mbox.\egroup }{2018}]{Yang}
Yang, X.; Wang, M.; Wang, W.; Khabsa, M.; and Awadallah, A.
\newblock 2018.
\newblock Adversarial training for community question answer selection based on
  multi-scale matching.
\newblock {\em arXiv preprint arXiv:1804.08058}.

\bibitem[\protect\citeauthoryear{Ying \bgroup et al\mbox.\egroup
  }{2018}]{Ying:GCN}
Ying, R.; He, R.; Chen, K.; Eksombatchai, P.; Hamilton, W.~L.; and Leskovec, J.
\newblock 2018.
\newblock Graph convolutional neural networks for web-scale recommender
  systems.
\newblock In {\em Proceedings of the 24th ACM SIGKDD International Conference
  on Knowledge Discovery; Data Mining},  974--983.
\newblock ACM.

\bibitem[\protect\citeauthoryear{Zafarani and Liu}{2009}]{zafarani2009social}
Zafarani, R., and Liu, H.
\newblock 2009.
\newblock Social computing data repository at asu.

\end{thebibliography}
}
\end{document}


\section {Appendix} \label{appendix}
\begin{proof}[Proof of Proposition 1]
According to the objective function in Eq. 2, we have:
\begin{align} \label{eq:V}
V(G,D)=\mathbb{E}_{x \sim p_+}(\log(D(x)))+\mathbb{E}_{x' \sim p_{-}^{g}}(\log(1-D(x)))
\end{align}
where $V(G,D)$ for a fixed $G$ achieves its maximum at $p_+/(p_{-}^{g}+p_+)$ since $f(y)=a\log(y)+b\log(1-y)$ (for any $(a,b) \in \mathbb{R}^2 \ (0,0)$) reaches its maximum (for $y \in [0,1]$) at $\frac {a}{a+b}$ (Goodfellow et al. 2014). 

\nonumber
\end{proof}

\begin{proof}[Proof of Theorem 1]
For the optimal discriminator, Eq. 2 can be reformulated as:
\begin{align} \label{eq:G_star}
min_{G}\frac{1}{|S^+|}\Sigma_{x \in S^+}\log{\frac{p_+(x)}{p_+(x)+p_{-}^{g}(x)}}+\\
\Sigma_{x \in S^-}G(x)\times \log{(1-\frac{p_+(x)}{p_+(x)+p_{-}^{g}(x)}}) +\\ \lambda \times \Sigma_{x \in S^-}p_{-}^{g}(x) \times \log(p_{-}^{g}(x)) \nonumber
\end{align}
If we set the gradient of the above cost function w.r.t. $G$ to zero, the optimal $G$ must satisfy the following equation:
\begin{align} \label{eq:G_star2}
\frac {(p_{-}^{g}(x))^{\lambda+1}}{\int (p_{-}^{g}(x))^{\lambda+1}dx}=\frac {p_{-}^{g}(x)+p_+(x)} {2}.
\end{align}
\end{proof}